\documentclass{IEEEtran}
\pdfoutput=1
\usepackage{makeidx}  
\usepackage{url}
\usepackage{hyperref}
\usepackage{pdfcomment}

\usepackage[cmex10]{amsmath}
\usepackage{amssymb}
\usepackage{latexsym}
\RequirePackage{bm}

\usepackage[tight,footnotesize]{subfigure}
\usepackage{graphicx}

\usepackage{eso-pic}
\usepackage{graphicx}

\newcommand{\mat}{\bm}
\renewcommand{\vec}[1]{\bm{\mathit #1}}
\newcommand{\etal}{\textit{et al.}}

\newcommand{\filter}{\mat I_{\mathrm f}}
\renewcommand{\binary}{\mat I_{\mathrm {thr}}}
\newcommand{\binaryA}{\mat I^\mathrm{A}_{\mathrm {thr}}}
\newcommand{\binaryB}{\mat I^\mathrm{B}_{\mathrm {thr}}}

\newcommand{\DSA}{\mat I_{\mathrm{DSA}}}
\newcommand{\contrasted}{\mat I_{\mathrm c}}
\newcommand{\uncontrasted}{\mat I_{\mathrm u}}
\newcommand{\sigmoid}{\mat I_\mathrm{sig}}

\newcommand{\shadowA}{\mat S^{\mathrm{A}}_T}
\newcommand{\shadowB}{\mat S^{\mathrm{B}}_T}

\newcommand{\DOG}{\mat I_\mathrm{DOG}}
\newcommand{\DOGA}{\mat I_\mathrm{DOG}^\mathrm{A}}
\newcommand{\DOGB}{\mat I_\mathrm{DOG}^\mathrm{B}}
\DeclareMathOperator*{\argmin}{arg\,min}
\DeclareMathOperator*{\argmax}{arg\,max}
\newcommand{\NCC}{\rho_\mathrm{n}}
\newcommand{\NCCCAD}{\rho_\mathrm{CADE}}
\newcommand{\NCCEDG}{\rho_{\mathrm{edge}}}
\newcommand{\NCCADSA}{\rho_{\mathrm{shad}}^{\mathrm{DSA}}}
\newcommand{\NCCATHR}{\rho_{\mathrm{shad}}^{\mathrm{thr}}}

\begin{document}
\include{responses}
\cleardoublepage
\twocolumn
\pagenumbering{arabic}
\setcounter{page}{1}
\title{3-D/2-D Registration of Cardiac Structures by 3-D Contrast Agent Distribution Estimation}

\author{Matthias~Hoffmann, Christopher~Kowalewski, Andreas~Maier, Klaus~Kurzidim, Norbert~Strobel, Joachim~Hornegger
\thanks{This work was supported by the German Federal Ministry of Education and Research (BMBF) in the context of the initiative Spitzencluster Medical Valley - Europ\"aische Metropolregion N\"urnberg, project grant Nos. 12EX1012A and 12EX1012E, respectively. Additional funding was provided by Siemens Healthcare GmbH.
J. Hornegger and A. Maier gratefully acknowledges funding of the Erlangen Graduate School in Advanced Optical Technologies (SAOT) by the German Research Foundation (DFG) in the framework of the German excellence initiative.
The concepts and information presented in this paper are based
on research and are not commercially available.} 
\thanks{M. Hoffmann, A. Maier and J. Hornegger are with Pattern Recognition Lab, Friedrich-Alexander-Universit\"at Erlangen-N\"urnberg, Martensstr. 3, 91058 Erlangen, Germany. E-mail: Matthias.Hoffmann@cs.fau.de}
\thanks{A. Maier and J. Hornegger are with Erlangen Graduate School in Advanced Optical Technologies (SAOT), Friedrich-Alexander-Universit\"at Erlangen-N\"urnberg, Paul-Gordan-Str. 6, 91058 Erlangen, Germany}
\thanks{C. Kowalewski and K. Kurzidim are with
Klinik f\"ur Herzrhythmusst\"orungen,	Krankenhaus Barmherzige Br\"uder Regensburg, Pr\"ufeninger Stra\ss e 86, 93049 Regensburg }
\thanks{
N. Strobel is with
Siemens Healthcare GmbH, Siemensstr. 1, 91301 Forchheim, Germany
}}

\maketitle              

\begin{abstract}
For augmented fluoroscopy during cardiac catheter ablation procedures, a preoperatively acquired 3-D model of the left atrium of the patient can be registered to X-ray images.
Therefore the 3D-model is matched with the contrast agent based appearance of the left atrium.
Commonly, only small amounts of contrast agent (CA) are used to locate the left atrium.
This is why we focus on robust registration methods that work also if the structure of interest is only partially contrasted.
In particular, we propose two similarity measures for CA-based registration: 
The first similarity measure, explicit apparent edges, focuses on edges of the patient anatomy made visible by contrast agent and can be computed quickly on the GPU. The second novel similarity measure computes a contrast agent distribution estimate (CADE) inside the 3-D model and rates its consistency with the CA seen in biplane fluoroscopic images.
As the CADE computation involves a reconstruction of CA in 3-D using the CA within the fluoroscopic images, it is slower. 
Using a combination of both methods, our evaluation on 11 well-contrasted clinical datasets yielded an error of 7.9$\pm$6.3\,mm over all frames. For 10 datasets with little CA, we obtained an error of 8.8$\pm$6.7\,mm.
Our new methods outperform a registration based on the projected shadow significantly ($p<0.05$).
\end{abstract}

\section{Introduction}
Atrial fibrillation is the most common heart arrhythmia
affecting around 2.2 million people in the US. 
A possible treatment option is catheter ablation, which is a minimally invasive procedure.
It is carried out using either electroanatomic mapping systems, a fluoroscopy guided approach or a combination of both.
In this paper, we refer to fluoroscopy guided approaches.
Unfortunately, X-ray images suffer from poor soft-tissue contrast such that the left atrium (LA) can only be seen if contrast agent (CA) is injected.
However, to reduce the risk of contrast-induced nephropathy, physicians try to keep the use of CA to a minimum often highlighting only a part of the left atrium.
To provide orientation to the physician when no CA is present, a model of the LA, e.g., generated by a CT or MRI scan of the patient can be overlaid~\cite{deBuck2005AugmentedReality}. 
As the coordinate systems of the preprocedurally acquired 3-D heart model and the patient during the intervention differ,
a registration step has to be performed. 
In clinical practice, this registration is usually carried out manually often involving a CA injection~\cite{bourier2012registration}.
Unfortunately, manual registration complicates the workflow. 
It either increases the workload of the treating physician, or it involves a trained assistant.
Therefore, automatic registration is preferred.

There has been much research about registration of 3-D objects to 2-D fluoroscopic images, e.g. for bones~\cite{Gueziec1998AnatomyBased,Hamadeh1998AutomatedRegistration} or implants~\cite{Kaptein2003RSA}. 
An overview is given by Markelj \etal~\cite{Markelj2012reviewregistration}.
Compared to implants, a registration of the LA is more complicated for two reasons: First, for implants and bones, usually the whole object is visible under fluoroscopy. 
During a contrast injection, however, only parts of the left atrium may be visible in the fluoroscopic images. 
Second, the general visibility of the LA may be poor depending on how much CA is used.
As a consequence, further effort is needed to develop robust registration methods that can also be applied if CA is used sparingly.

In a first approach towards automatic LA registration, Thivierge-Gaulin \etal~\cite{thivierge2012registration} tried to find a 3-D pose of a model such that its projected shadow matches the contrasted area in a selected image, enhanced by digital subtraction angiography (DSA), best.
Based on CT images, a second approach by Zhao \etal~\cite{Zhao2013Registration} relied on digitally rendered radiographies of the segmented left atrium.
The rendered image was compared to a DSA image using normalized gradient correlation where distinct regions of the atrium were weighted differently.


We propose two new registration techniques for contrast agent-based registration:
In our first method, we take explicitly apparent edges extracted from a 3-D model and compare them to LA edges present in the fluoroscopic images as proposed by~\cite{Gueziec1998AnatomyBased,Hamadeh1998AutomatedRegistration} for automatic registration of bones and by \cite{hoffmann2013Visualization} for manual registration of the LA, respectively.
This comparison can be carried out quickly on a GPU.
Second, we introduce a novel similarity measure for biplane fluoroscopy that is tailored for cases in which only parts of an object are visible.
Based on a 3-D model of the LA, our second method estimates the contrast agent distribution inside the 3-D object from a simultaneously acquired pair of fluoroscopic images taken under two different view angles.
Then we evaluate how consistent the contrast agent distribution estimate (CADE) is with the acquired fluoroscopic images.
As the CADE depends on the transformation used for registration, the transformation leading to the most plausible CADE is used as final position estimate.

\begin{figure}[t]%
\begin{center}
\includegraphics[width=0.7\linewidth]{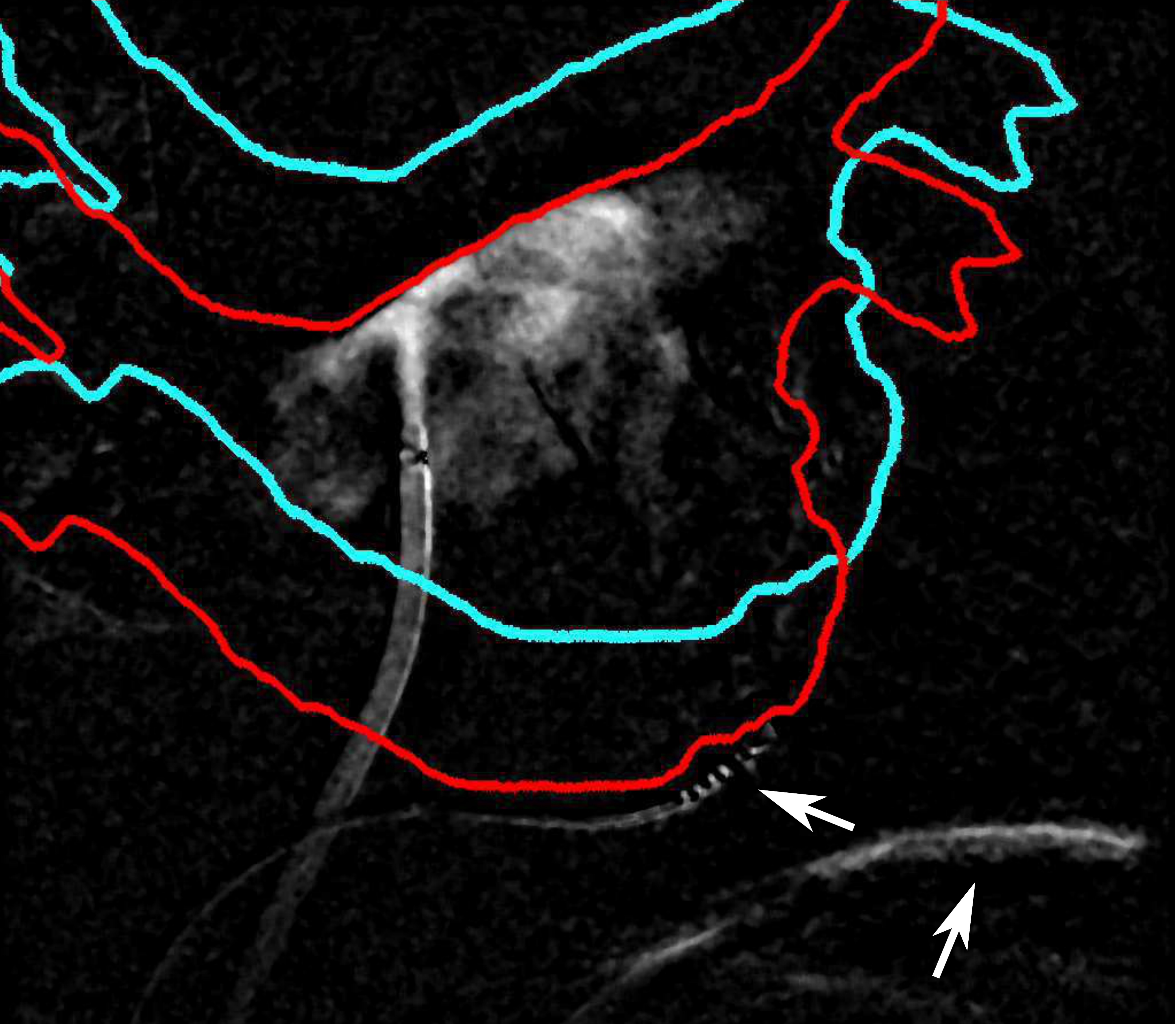}
\end{center}
\caption{A correct (red) and wrong (cyan) registration result. 
In both cases, the contrasted area is fully inside the projection shadow represented by the colored outline.
This leads to a similar NCC value when using only an area-based feature for automatic registration. 
Thanks to the best reference frame selection, motion artifacts could be kept to a minimum. Only some remained in the vicinity of the moving coronary sinus (CS) catheter and the diaphragm, see white arrows.}
\label{fig:objfktarea}
\end{figure}
\section{Registration Method}\label{sec:method}
For registration, two fluoroscopic sequences showing a CA injection are used.
These sequences are acquired simultaneously from two different angles using an angiography biplane system.
For each plane, the projection matrix that describes the X-ray camera setup is known.
We denote the associated projection operator by $P$.
We also assume that a 3-D model of the patient's LA is available, either as a triangle mesh or a binary volume, as they can be converted into each other.

\subsection{Contrast Agent Extraction}\label{ssec:contrastExtraction}
The contrasted area is found based on a difference image (DSA) $\DSA=\uncontrasted-\contrasted$, $\mat I\in\mathbb R^{m\times n}$ involving a frame $\contrasted$ that contains contrast agent and an uncontrasted frame $\uncontrasted$.
To distinguish between contrasted and uncontrasted frames, either manual annotation, a threshold based method e.g. the method described in~\cite{Zhao2013Registration} or an automatic contrast detection~\cite{Hoffmann2015ContrastDetection} can be used.
Depending on the chosen contrasted frame $\contrasted$, $\DSA$ may contain artifacts due to motion of the diaphragm or from catheters if they are at different positions in $\uncontrasted$ and $\contrasted$.
Such motion artifacts depend, unlike the information about contrast agent, to a large degree on the choice of $\uncontrasted$.
For example, if the catheters in $\uncontrasted$ are at the same position as in $\contrasted$, their intensities cancel out in the subtraction image.
Otherwise, $\DSA$ has high positive values at the position of the catheter in $\contrasted$ and high negative values at the position of the catheter in $\uncontrasted$.
To keep motion artifacts to a minimum, we propose a best reference selection, which chooses an appropriate reference frame $\hat\uncontrasted$ that matches the chosen contrasted frame $\contrasted$ as much as possible.
Out of all uncontrasted frames, that frame $\hat\uncontrasted$ is selected which minimizes the $L_1$-norm of the resulting DSA image
\begin{equation}\label{eq:contrast}
\hat\uncontrasted=\argmin_{\uncontrasted} \sum_{x=0}^n\sum_{y=0}^m |\uncontrasted(x,y)-\contrasted(x,y)|.
\end{equation}
By following Eq.~\ref{eq:contrast}, we get frames for which the catheters and the diaphragm cancel out as much as possible.
See Figure~\ref{fig:dsa} for an example.
In $\DSA$, only pixels with positive values contain contrast agent. To extract them, we set the intensity of pixels with negative value to 0.
Afterwards, we compute a filtered image $\filter$ by applying a median filter with a large kernel size.
Smaller structures, e.g., caused by motion artifacts that remained despite the optimized choice of $\uncontrasted$, do not pass this filter,
and the noise in the contrasted area is reduced as well.
Finally, a binary image $\binary$ of the filtered image $\filter$ is computed using a threshold at $\mu_\mathrm{f}+\sigma_\mathrm{f}$ where $\mu_\mathrm{f}$ and $\sigma_\mathrm{f}$ denote the mean and standard deviation of $\filter$, respectively. Thus, a contrasted pixel $\vec p \in \mathbb R^2$ is indicated by $\binary(\vec p) = 1$.

A previous approach~\cite{thivierge2012registration} tried to find a transformation $T$ of the 3-D model such that the projected shadows $\shadowA,\shadowB$ of the model into the A-plane and the B-plane of a biplane C-arm system fit best to the contrasted region.
Using the normalized cross correlation (NCC), denoted as $\NCC$, of two images $\mat I_1,\mat I_2$ with corresponding mean values $\mu_1,\mu_2$ and standard deviations $\sigma_1,\sigma_2$
\begin{equation}
\NCC(\mat I_1,\mat I_2)=\sum_{x=0}^n\sum_{y=0}^m \frac{(\mat I_1(x,y)-\mu_1)\cdot(\mat I_2(x,y)-\mu_2)}{\sigma_1\cdot\sigma_2},
\end{equation}
the similarity of the projected shadow and $\DSA$ can be measured. A registration transformation can be estimated by maximizing either one of the two functions
\begin{eqnarray}
\NCCADSA(T)=&\NCC(\DSA^{\mathrm A},\shadowA)&\cdot\NCC(\DSA^{\mathrm B},\shadowB),\\
\NCCATHR(T)=&\NCC(\binaryA,\shadowA)&\cdot\NCC(\binaryB,\shadowB).
\end{eqnarray}

\subsection{Edge Feature}\label{ssec:edge}

\begin{figure}[tb]%
\begin{center}

\subfigure[]{
 \includegraphics[width=0.45\linewidth]{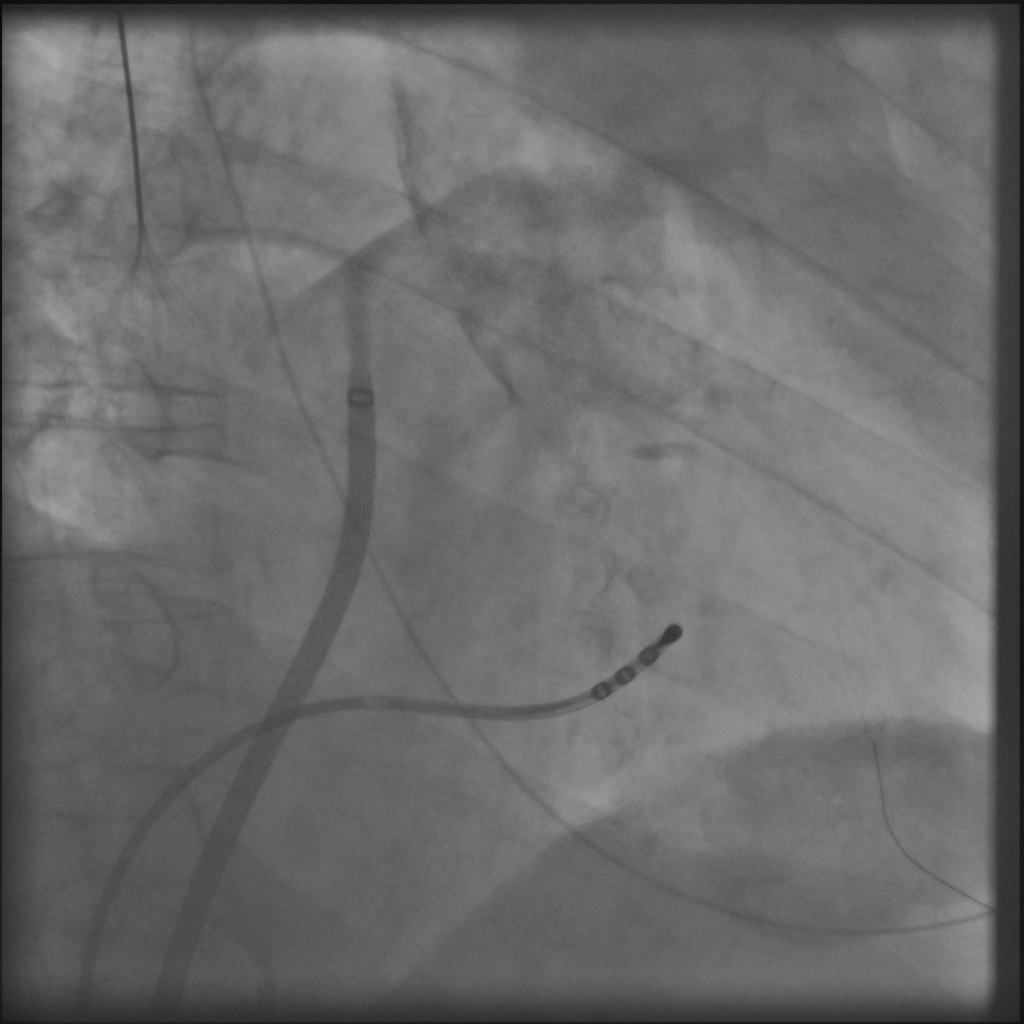}
 \label{fig:orig}
}~~
\subfigure[]{
 \includegraphics[width=0.45\linewidth]{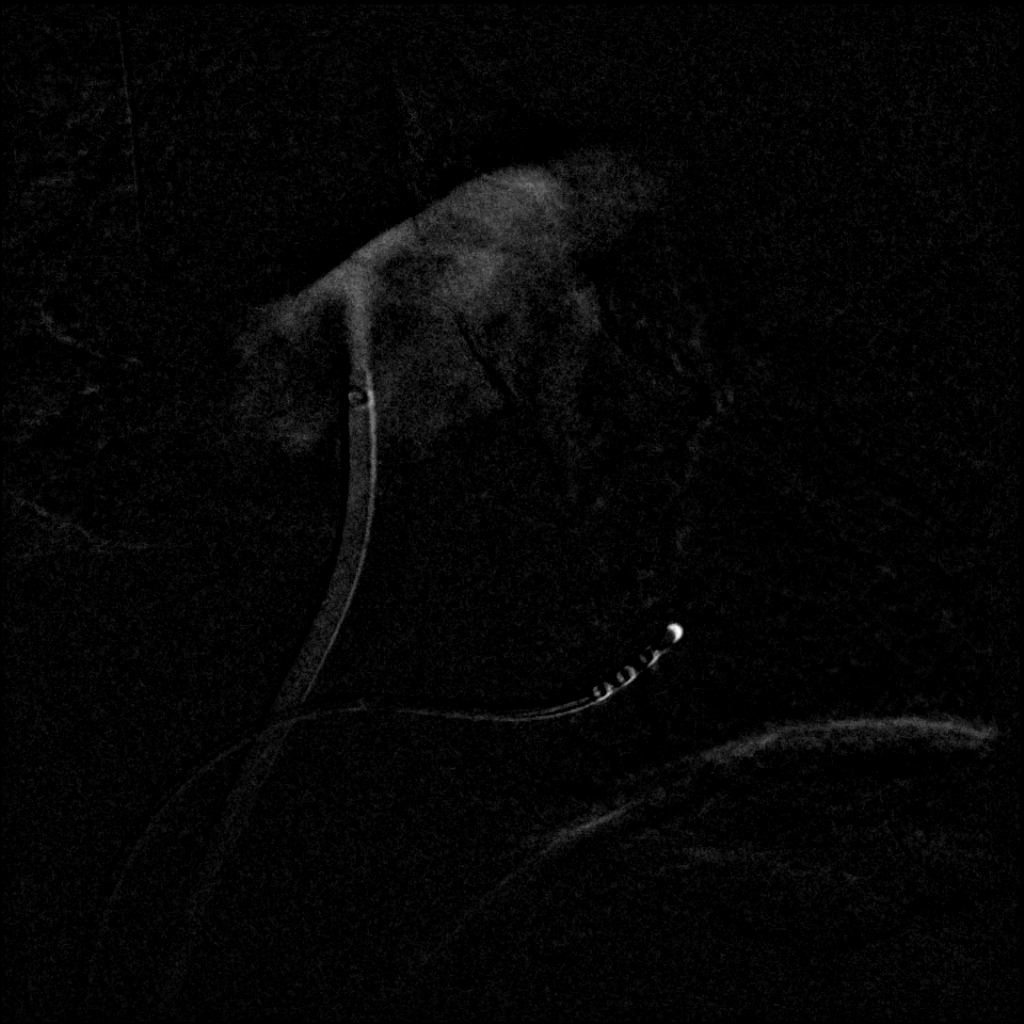}
 \label{fig:dsa}
}\quad
\subfigure[]{
 \includegraphics[width=0.45\linewidth]{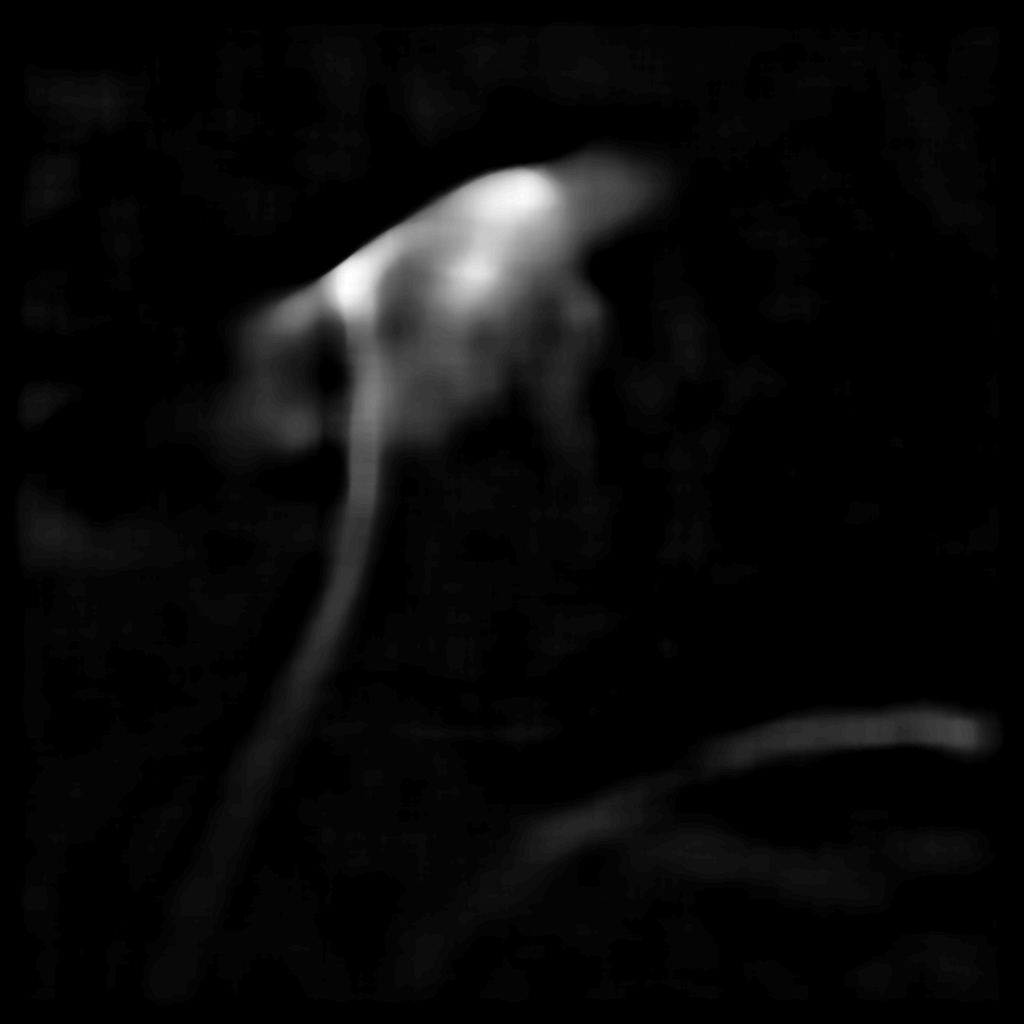}
 \label{fig:median}
}~~
\subfigure[]{
 \includegraphics[width=0.45\linewidth]{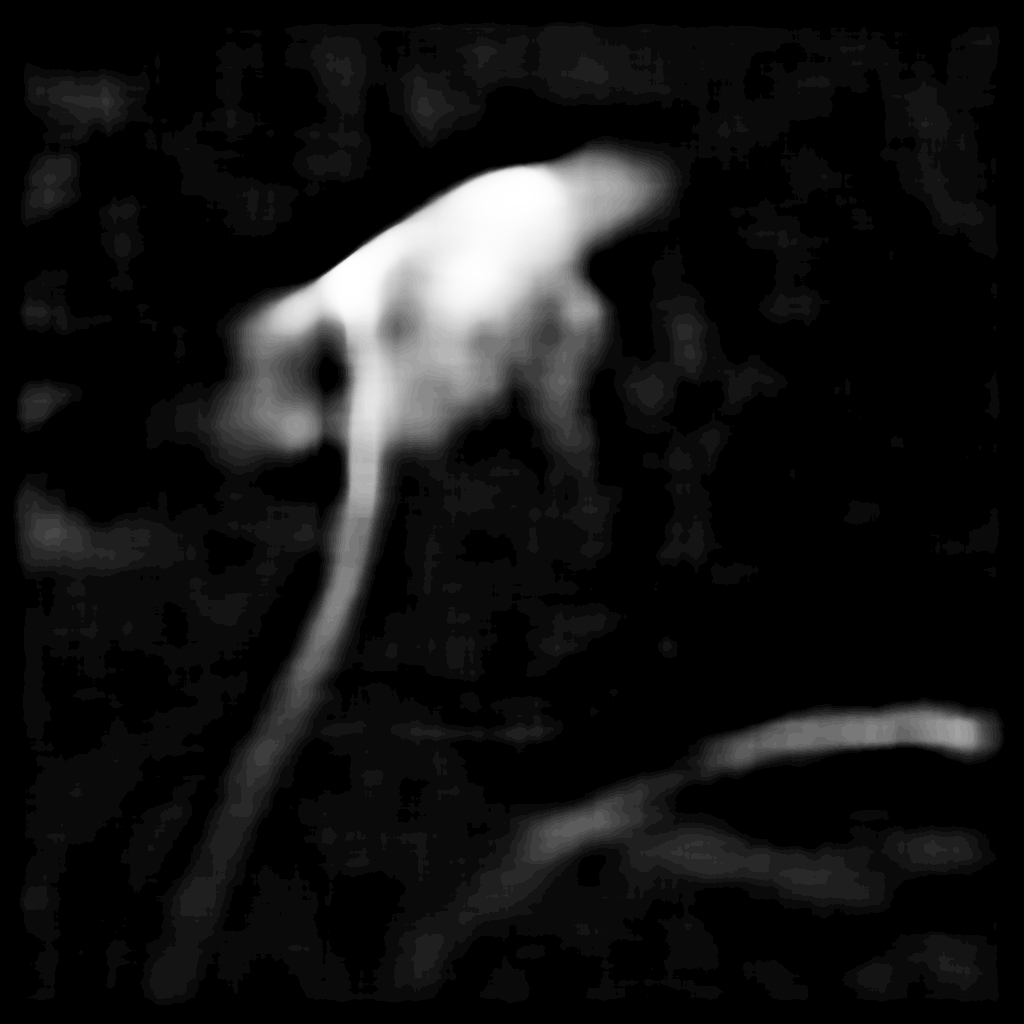}
 \label{fig:sigmoid}
}\quad
\subfigure[]{
 \includegraphics[width=0.45\linewidth]{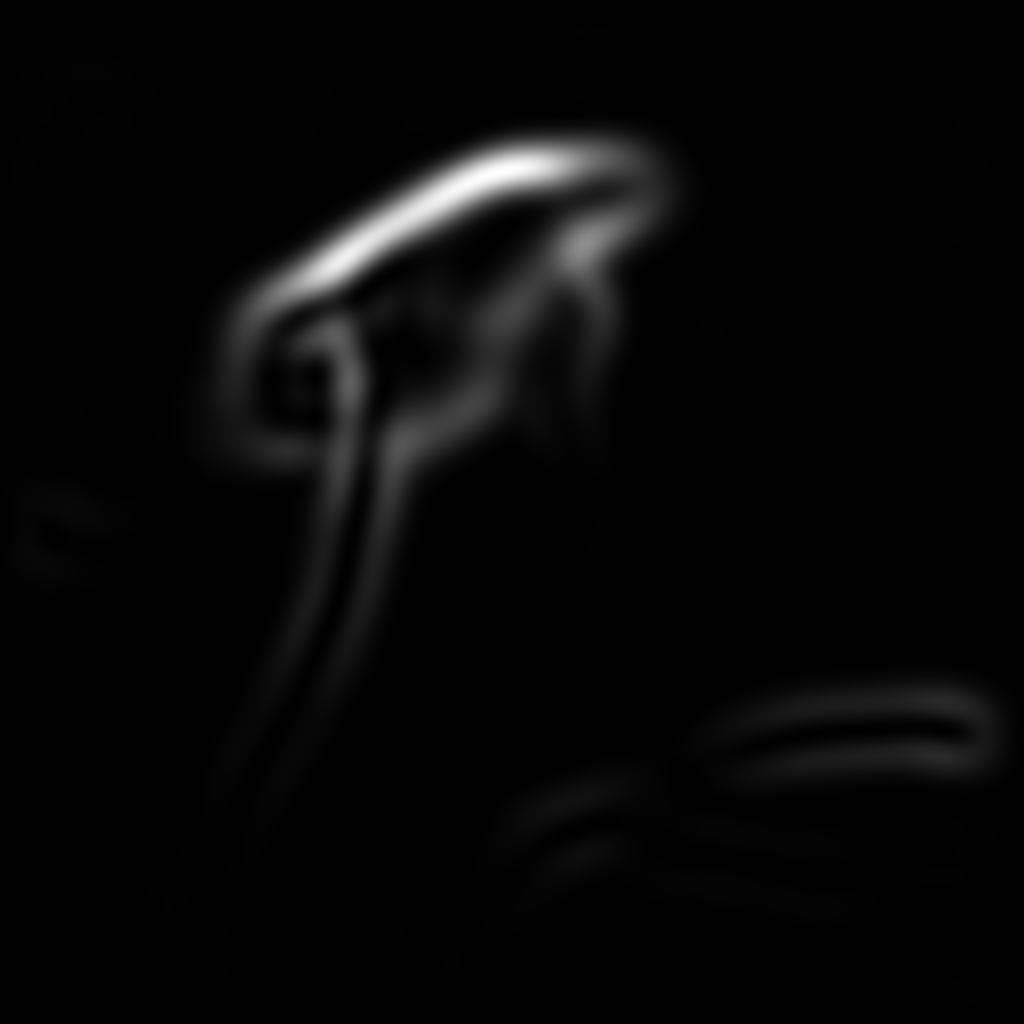}
 \label{fig:dog_large}%
}~~
\subfigure[]{
 \includegraphics[width=0.45\linewidth]{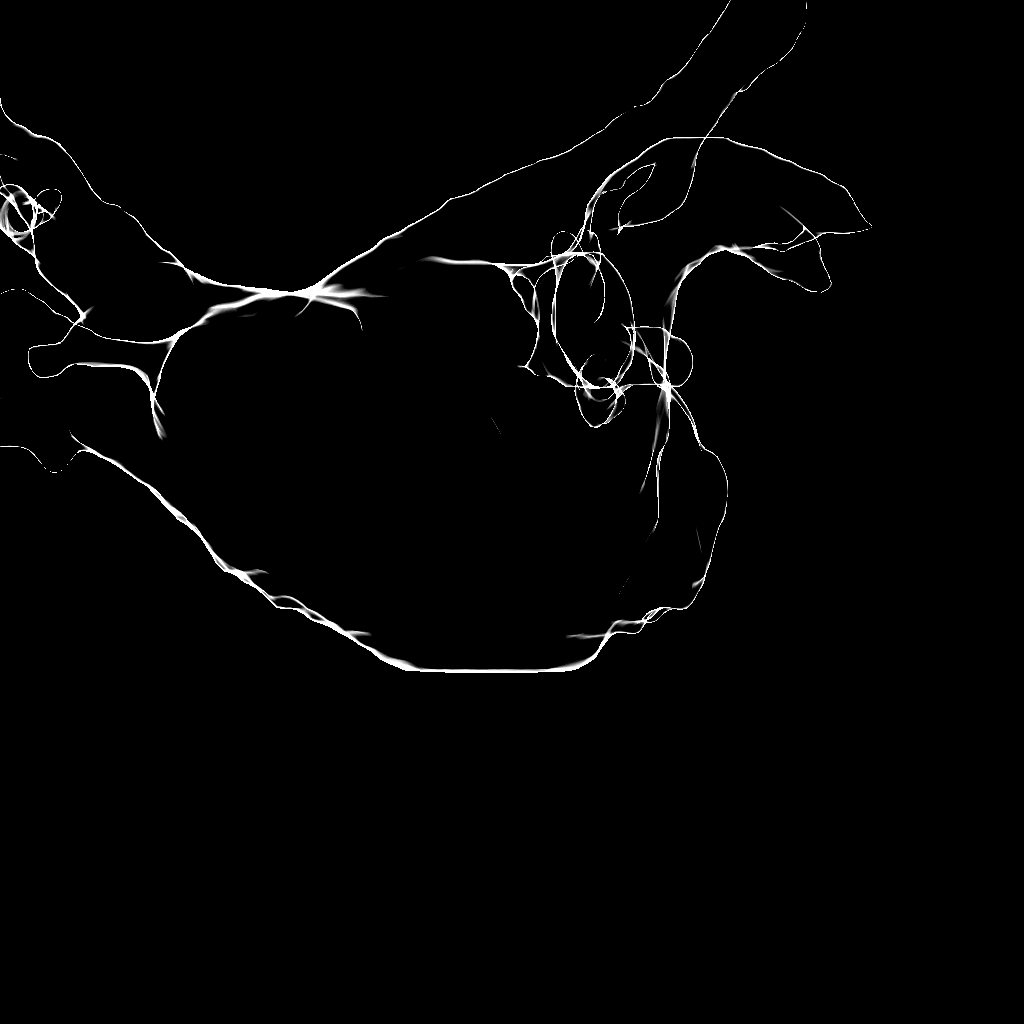}
 \label{fig:rendering}%
}
\end{center}
\caption{Using the original image (a), a DSA image (b) is computed. 
After median filtering (c), the remaining motion artifacts from the catheter have vanished. 
Afterwards, all pixels are weighted by a sigmoid function (d) to get a homogeneous value distribution inside the contrasted area.
Edges are extracted using derivatives of Gaussian with a large kernel size (e).
Finally, the similarity to the rendered edges (f) is evaluated. }%
\label{fig:overview}
\end{figure}

Unfortunately, a registration approach only based on contrasted area has multiple solutions if the amount of CA is so little that it can be located at different positions inside the LA, see \figurename~\ref{fig:objfktarea}.
Often, CA is injected against the roof or into the pulmonary veins.
This results in perceivable edges of the contrasted area which can be used as registration features as well.
Edge-based registration can be done using only the silhouette boundary of the projected object~\cite{Kaptein2003RSA}, see \figurename~\ref{fig:objfktarea} or all apparent edges~\cite{Gueziec1998AnatomyBased,Hamadeh1998AutomatedRegistration}, see \figurename~\ref{fig:rendering}.

We decided to go for the second approach, as the silhouette corresponds to the edges in the image only if the complete atrium is filled with contrast.
For a partially contrasted left atrium, internal contours may, however, also appear in the fluoroscopic images. 
This was already found to be beneficial for manual LA registration~\cite{hoffmann2013Visualization}.
Instead of considering edges implicitly by comparing the DSA image to a DRR using gradient correlation~\cite{Zhao2013Registration}, we computed them explicitly.
To extract edges in the fluoroscopic images, we used the previously computed filtered image $\filter$.
After applying a median filter, edge-like variations \emph{inside} the contrasted areas may remain. They would trigger a response, if an edge filter was applied. 
To obtain an edge response only at the boundaries of the contrasted area, the image needs to be homogenized before applying an edge filter.
Using a simple threshold method would result in a loss of the intensity drop-off at the boundary which provides important information about the edge intensity.
Therefore, we weigh all image pixels by a sigmoid function
\begin{equation}\label{eq:sig}
	\mat I_\mathrm{sig}(x,y) = \frac{1}{1 + \mathrm e^{-(\mat I_\mathrm{f}(x,y)+t)\cdot s}}.
\end{equation}
The value of $t$ is set to $\mu_\mathrm{f}-\sigma_\mathrm{f}$, and the parameter $s$ depends on the pixel intensity range of the input image.
An example of $\sigmoid$ is given in Figure~\ref{fig:sigmoid}.
Finally, $\sigmoid$ is filtered using a derivative of Gaussian (DOG) filter to obtain the edge image $\DOG$.
The kernel size of the DOG-filter is set to a large value to get a smooth similarity measure, see Figure~\ref{fig:dog_large}.

The projection of the 3-D triangle mesh edges into 2-D is done differently than in~\cite{Gueziec1998AnatomyBased,hoffmann2013Visualization}.
We rendered the whole surface mesh and, depending on the viewing direction $\vec d$ and the surface normal $\vec n$ at a point, we set the opacity of projected triangles to $o=1-(\vec d \circ \vec n)$, see Figure~\ref{fig:rendering} for an example.
By doing so, areas that are parallel to the imaging plane are rendered transparent while areas with a normal vector orthogonal to the viewing direction are rendered opaque.
The similarity between edges extracted from the fluoroscopic images and the edge images $\vec E^{\mathrm{A}}_{T},\vec E^{\mathrm{B}}_{T}$ rendered from the 3-D model transformed by $T$ is measured by
\begin{equation}
\NCCEDG(T)=\NCC(\DOGA,\vec E^{\mathrm{A}}_{T}) \cdot\NCC(\DOGB,\vec E^{\mathrm{B}}_{T}).
\end{equation}

\subsection{Contrast Agent Distribution Estimation (CADE)}\label{ssec:CADE}
Previous approaches~\cite{thivierge2012registration,Zhao2013Registration} for LA registration searched for a rigid transformation of the LA such that either its projected shadow or its DRR fit to the contrasted area in both fluoroscopic images.
3-D information was integrated insofar as the resulting projections came from the same 3-D position of the model.
Unfortunately, such an approach does not guarantee that corresponding objects in both fluoroscopic images are matched to the same 3-D structure of the LA.
More precisely, the registration result could be such that in plane A, the contrast agent is located in a left pulmonary vein (PV) whereas in plane B, the contrasted area corresponds to a right PV.
This is possible as for a given 2-D registration in one plane, the 2-D registration in the other plane has one degree of freedom, which corresponds to an out-of-plane motion in the first plane.

To solve this problem, we compute for a given transformation $T$ a CADE inside the LA using binary reconstruction. 
Then, $T$ is optimized such that the contrast agent distribution estimate is most consistent with the projection images.
More precicely, a voxel $\vec v$ is estimated as contrasted if it fulfills all of the following conditions:
The voxel $\vec v$ transformed by $T$ is (a) projected on a contrasted pixel in plane A, (b) projected on a contrasted pixel in plane B, and (c) $\vec v$ is part of the left atrium as contrast agent can only be found inside the left atrium.
For the computation of the CADE, we define therefore the indicator function
\begin{equation}
\chi(\vec v)=1 \Leftrightarrow \vec v \in \mathbb R^3\text{ is inside the left atrium}.
\end{equation}
Given the binary images $\binaryA$ and $\binaryB$ with corresponding projection operators $P_A, P_B$ and the indicator function $\chi(\vec v)$, the CADE $C^{\mathrm{3-D}}_T$ can be computed as
\begin{equation}
C^{\mathrm{3-D}}_T(\vec v) = \binaryA\left(P_A\left(T(\vec v)\right)\right) \cdot \binaryB(P_B(T(\vec v))) \cdot \chi(\vec v)
\label{eq:CADE}
\end{equation}
for a given rigid transformation $T$.
These three factors correspond to the aforementioned conditions.

\begin{figure}[tb]%
\begin{center}
 \includegraphics[width=0.9\linewidth]{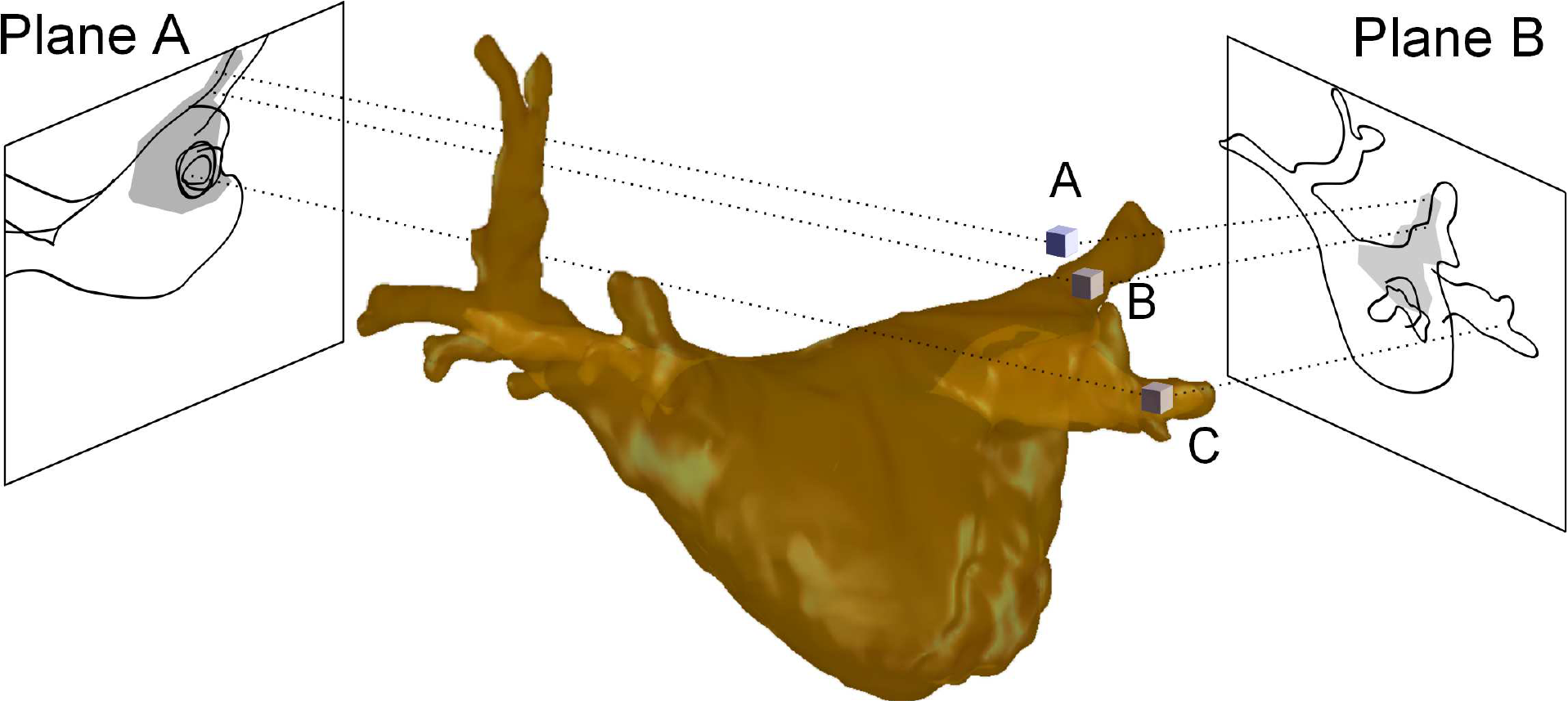}
\end{center}
\caption{For the transformation shown, only voxel B is estimated as containing CA as it is inside the LA and contrasted in both planes. 
Voxel C is only contrasted in Plane A but not in Plane B. Voxel A is filled with contrast in both planes, but it is outside of the LA and therefore considered as uncontrasted.
If the LA is moved such that its projections in A and B cover the contrasted area, this voxel will also be estimated as contrasted.}%
\label{fig:CAD}
\end{figure}
If $T$ is chosen suboptimally, the resulting 3-D CADE will be inconsistent with the CA observed in the 2-D images.
I.e. a pixel in the 2-D image is constrasted but no corresponding voxel along its projection ray is estimated as contrasted.
This can be due to following reasons as shown in Figure~\ref{fig:CAD}:
	(a) the projection ray from a contrasted pixel does not intersect the left atrium as the LA has not been placed at the proper position yet;
	(b) the projection ray hits the LA, but all voxels intersected by this ray cannot contain CA because their corresponding pixels in the other plane are uncontrasted.
Additional inconsistencies are introduced by pixels which are erroneously labeled as contrasted e.g. due to motion artifacts.
To verify the validity of the CADE, we compute 2-D images $\vec C^\mathrm{A}_{T}, \vec C^\mathrm{B}_{T}$ by forward projecting all contrasted voxels in $C^{\mathrm{3-D}}_T$ using $P_A, P_B$.
We assess the consistency of the CADE for the given transformation $T$ by computing the similarity between the fluoroscopic images and the projected CADE by
\begin{equation}
\NCCCAD(T)=\NCC(\binaryA,\vec C^\mathrm{A}_{T}) \cdot\NCC(\binaryB,\vec C^\mathrm{B}_{T}).
\end{equation}

\section{Experiments and Results}\label{sec:experiments}
We evaluated our method on 21 clinical biplane X-ray sequences from 10 different patients.
The data set contained 11 sequences showing an initial contrast agent injection where CA was injected into the LA center through a sheath. 
Besides the sheath, only the coronary sinus catheter was present. 
There were 10 more sequences showing a secondary injection for re-registration. 
Here, less CA was injected and additional catheters were present.
For all 133 contrasted frames, reference registrations performed by three clinical experts were available.
These reference registrations covered only translation as in~\cite{bourier2012registration}.

As initialization for optimization, the 3-D model was placed at that 3-D position which corresponded to the centers of both 2-D images.
In some cases, the initialization was more than 30\,mm away from the correct solution and beyond the capture range for gradient-based methods.
Therefore we applied an octree-like coarse-to-fine scheme where we evaluated several positions at a coarse resolution.
At positions in space that yielded a good similarity value, we performed subsequent evaluations on an increasingly finer resolution.
The 3-D translation $\hat {\vec t}=\argmax_{\vec t}\rho({\vec t})$ found by the optimization process of the respective objective function $\rho$ was compared to the mean translation vector ${\vec t}^*$ of the three manual registration results.
The distance $||\hat {\vec t} - {\vec t}^*||_2$ was used as error measure.
The significance of the results was measured using a Wilcoxon signed-rank test and a significance level of $0.05$.

All sequences contained 12-bit images of size $1024\times1024$\,pixels, all image processing, including rendering from the 3-D model, was performed on the full image size. 
The kernel size for median filter was 30\,pixels, the value $s$ of Eq.~\ref{eq:sig} was 0.1. The standard deviation of the DOG filter was 24\,pixels.

In the evaluation, we compared the similarity measures $\NCCATHR$, $\NCCADSA$, $\NCCCAD$ and $\NCCEDG$ and 
the combined similarity measures $\NCCATHR$+$\NCCEDG$, $\NCCADSA$+$\NCCEDG$ and $\NCCCAD$+$\NCCEDG$.
We investigated different weightings. Giving both terms equal weights turned out to be a good choice.
During evaluation, a registration for all frames marked as contrasted was performed.

\begin{table}[tbp]
\begin{center}
\caption{Translation errors for all frames}
    \begin{tabular}{l|ll}
    Objective function 	& initial injection				& subsequent injections				\\
		\hline
    $\NCCADSA$					&9.3$\pm$6.9\,mm    			&9.8$\pm$4.8\,mm						\\
    $\NCCATHR$					&9.9$\pm$7.9\,mm    			&12.1$\pm$9.1\,mm   				\\
		$\NCCEDG$					  &12.0$\pm$8.9\,mm    		&14.6$\pm$8.7\,mm   				\\
		$\NCCCAD$						&\textbf{8.7$\pm$6.4\,mm}	&\textbf{9.0$\pm$5.9\,mm } \\
		\hline
    $\NCCADSA+\NCCEDG$	&8.3$\pm$6.6\,mm     			&9.6$\pm$5.4\,mm    				\\
		$\NCCATHR+\NCCEDG$	&8.3$\pm$6.7\,mm     			&10.5$\pm$7.6\,mm   				\\
		$\NCCCAD+\NCCEDG$		&\textbf{7.9$\pm$6.3\,mm} 		&\textbf{8.8$\pm$6.7\,mm}  \\
		\hline
		Clinical experts		&3.3$\pm$2.7\,mm &3.1$\pm$1.7\,mm  \\
		\end{tabular}
		\label{tab:allFrames}
		\end{center}
\end{table}%
We computed a registration for each contrasted frames and compared the result to the manual registration of the physicians.
These results are clinically relevant if the physician requires a registration for a frame he determines, e.g depending on the breathing phase.
As potentially every frame could be selected by the physician, the overall accuracy should be high.
The overall accuracy is also of importance if the results are post-processed, e.g. a temporal filtering is applied.

\begin{figure}[tb]%
\begin{center}

\subfigure[]{
 \includegraphics[width=0.3\linewidth]{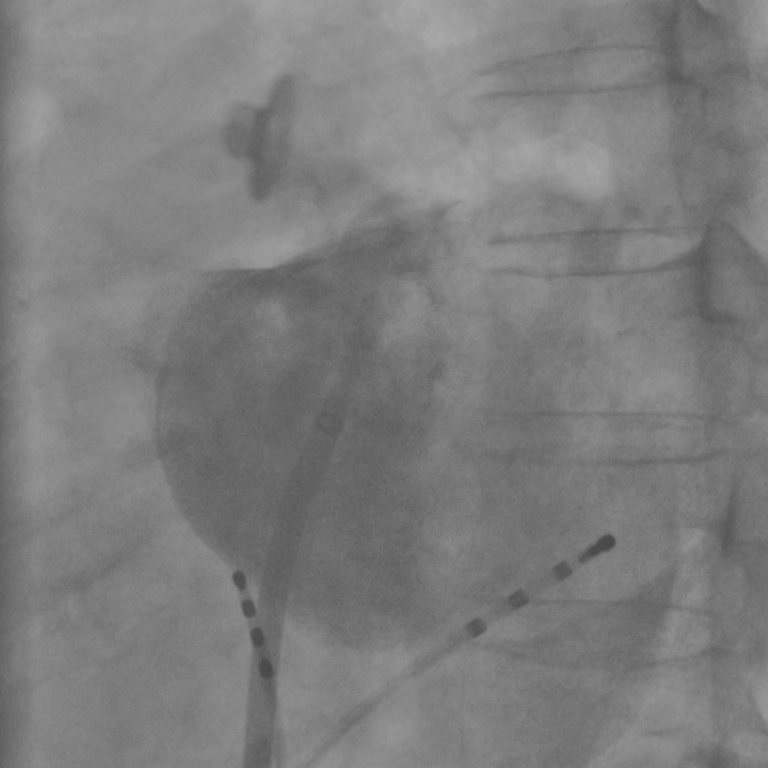}
 \label{fig:resoirg}
}~
\subfigure[]{
 \includegraphics[width=0.3\linewidth]{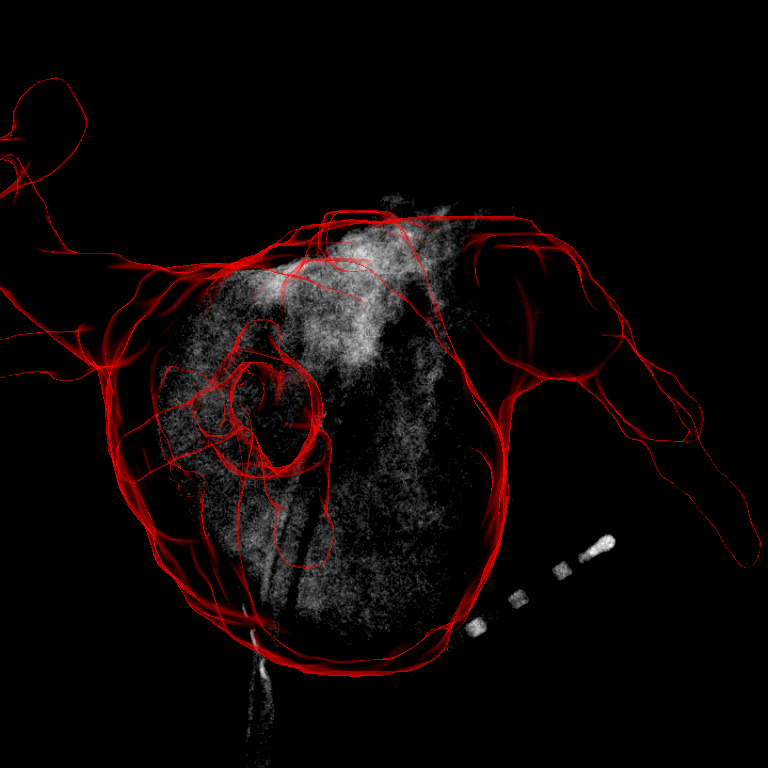}
 \label{fig:resarea}
}~
\subfigure[]{
 \includegraphics[width=0.3\linewidth]{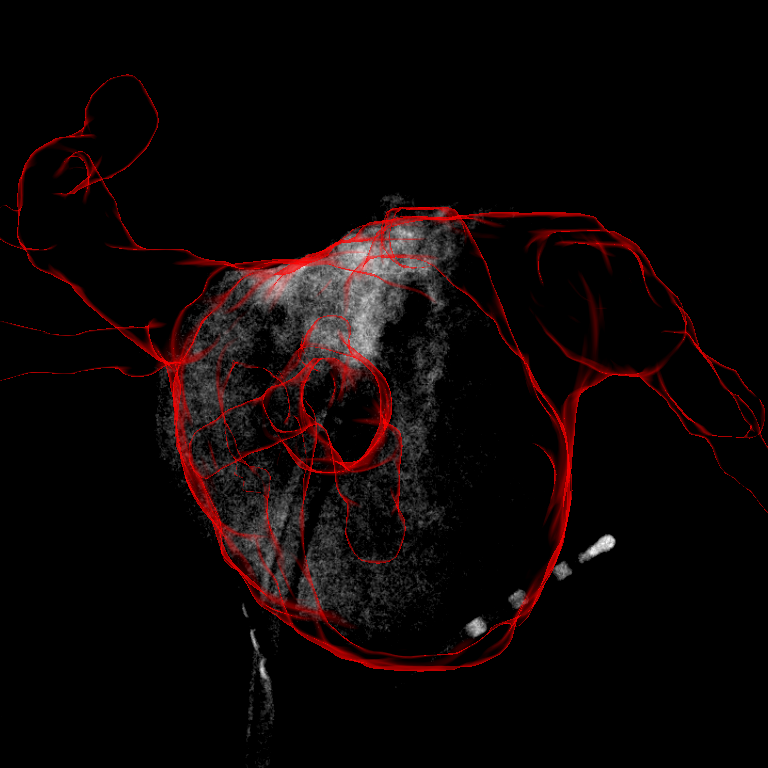}
 \label{fig:rescade}
}~
\end{center}
\caption{(a) Contrasted fluoroscopic image. The registration result when using $\NCCADSA$ (b) had an error of 6.7\,mm. By using $\NCCCAD$+$\NCCEDG$ (c), the left border of the LA model fits better to the left edge of the CA and the error reduced to 3.1\,mm }%
\label{fig:qualiResults}
\end{figure}

The evaluation results are presented in \tablename~\ref{tab:allFrames}.
The overall inter-user-variablity observed in the manual registrations was 3.2$\pm$2.3\,mm.
Considering all sequences, $\NCCATHR$ and $\NCCADSA$ gave significant better results when they are combined with $\NCCEDG$.
Also the performance of $\NCCCAD+\NCCEDG$ was significantly better than $\NCCATHR$, $\NCCADSA$ and $\NCCATHR+\NCCEDG$.
Compared to other measures, $\NCCEDG$ gave significantly worse results.
An example for a result is given in \figurename~\ref{fig:qualiResults}.

The image preprocessing takes 0.5\,s on an Intel Xeon E3 with 3.4 GHz and 16\,GB RAM.
The evaluations of the similarity measures were performed completely on the GPU.
On an NVIDIA GeForce GTX 660 the evaluation of $\NCCADSA$, $\NCCATHR$ and $\NCCEDG$ took 1.8\,ms for a given translation and 13.4$\pm$3.7\,ms for $\NCCCAD$.
The whole registration for a single frame takes 2.9\,s for $\NCCADSA$ and $\NCCATHR$ and 21.5$\pm$5.9\,s for $\NCCCAD$.
For a combination with $\NCCEDG$ it takes 5.8\,s and 27.1$\pm$5.9\,s, respectively.

\section{Discussion and Conclusions}\label{sec:discussion}
Our novel CADE based method outperformed the shadow based similarity measures $\NCCADSA$ and $\NCCATHR$, especially for re-registration sequences where only a small amount of contrast agent was used.
For these sequences, registration errors leading to inconsistent results could be avoided by $\NCCCAD$.
For well contrasted sequences, the improvement by $\NCCCAD$ is less as inconsistencies play a minor role.
We found that the similarity measure using explicit apparent edges, $\NCCEDG$, yields poor results when used by its own, but can improve results significantly when combined with other similarity measures.

It remains open if the accuracy for all frames, which is between between 7\,mm and 9\,mm, is sufficient for a clinical application.
In a future work, temporal constraints on the heart movement could be included in the registration process or a temporal filtering could be applied afterwards.

Compared to the approach by Zhao~\etal~\cite{Zhao2013Registration}, the determination of special weightings for different heart regions, and, for $\NCCATHR$+$\NCCEDG$, a time consuming DRR generation was avoided.
To summarize, a registration based on a combination of shadow and edge features leads to a registration that can be computed fast.
If it is possible to use more time for registration, the novel CADE-based measure, which estimates consistency, should be used as it leads to better results.

\bibliographystyle{abbrv}
\bibliography{literature}
\end{document}